\documentclass[letterpaper, 10 pt, conference]{ieeeconf}  

\IEEEoverridecommandlockouts                              

\usepackage{graphics} 
\usepackage{epsfig} 
\usepackage{amsmath, amsfonts} 
\usepackage{amssymb}  
\usepackage{bm}
\usepackage{siunitx}
\usepackage{subcaption}
\usepackage{algorithmicx}
\usepackage{algorithm}%
\usepackage{algorithmicx}%
\usepackage{algpseudocode}%
\usepackage{stfloats}

\usepackage[font={small}]{caption}
\usepackage{cite}
\usepackage{tikz}
\usetikzlibrary{automata, positioning, arrows.meta, shapes.multipart}

\usepackage[nolist,nohyperlinks]{acronym}
\acrodef{GPS}[GPS]{Global Positioning System}
\acrodef{SLAM}[SLAM]{Simultaneous Localization And Mapping}
\acrodef{GPS}[GPS]{Global Positioning System}
\acrodef{RTK}[RTK]{Real-time Kinematics}
\acrodef{GNSS}[GNSS]{Global Navigation Satellite System}
\acrodef{ROS}[ROS]{Robot Operating System}
\acrodef{UAV}[UAV]{Unmanned Aerial Vehicle}
\acrodef{IMU}[IMU]{Inertial Measurement Unit}
\acrodef{LiDAR}[LiDAR]{Light Detection and Ranging}
\acrodef{LKF}[LKF]{Linear Kalman Filter}
\acrodef{EKF}[EKF]{Extended Kalman Filter}
\acrodef{IEEE}[IEEE]{Institute of Electrical and Electronics Engineers}
\acrodef{MRS}[MRS]{Multi-robot Systems Group}
\acrodef{AGL}[AGL]{Above Ground Level}
\acrodef{DEM}[DEM]{Digital elevation model}
\acrodef{VIO}[VIO]{Visual-Inertial Odometry}
\acrodef{LIO}[LIO]{Lidar-Inertial Odometry}
\acrodef{RMSE}[RMSE]{Root Mean Square Error}
\acrodef{AMSL}[AMSL]{Above Mean Sea Level}

\newcommand{\appropto}{\mathrel{\vcenter{
  \offinterlineskip\halign{\hfil$##$\cr
    \propto\cr\noalign{\kern2pt}\sim\cr\noalign{\kern-2pt}}}}} 
\newcommand{\reffig}[1]{Figure~\ref{#1}}

\newcommand{\refsec}[1]{Section~\ref{#1}}
\newcommand{\reftab}[1]{Table~\ref{#1}}
\usepackage{siunitx}
\DeclareSIUnit \parsec {pc}
\DeclareSIUnit \electronvolt {eV}
\DeclareSIUnit \pixel {px}
\DeclareSIUnit \arcmin {arcmin}
\DeclareSIUnit \erg {erg}
\DeclareSIUnit \joul {J}

\tikzset{
    image_label/.style={
        fill=white,
        fill opacity=1.0, 
        text opacity=1,
        font=\bfseries,
        inner sep=3pt,
        minimum size=18pt,
        rectangle
    }
}
\usepackage{url}

\usepackage[breaklinks]{hyperref}

\usepackage[T1]{fontenc}
\usepackage{newtxtext,newtxmath}

\usepackage{booktabs}   

\begin{document}

\title{\LARGE \bf
Kilometer-Scale GNSS-Denied UAV Navigation via Heightmap Gradients: A Winning System from the SPRIN-D Challenge
}

\author{Michal Werner, David \v{C}apek, Tom\'{a}\v{s} Musil, Ond\v{r}ej Franek, Tom\'{a}\v{s} B\'{a}\v{c}a, and Martin Saska\\
\small Faculty of Electrical Engineering, Czech Technical University in Prague
\\
\small {\tt\footnotesize wernemic@fel.cvut.cz}
\vspace{-1em}
}

\maketitle
\thispagestyle{empty}
\pagestyle{empty}

\begin{abstract}
Reliable long-range flight of unmanned aerial vehicles (UAVs) in GNSS-denied environments is challenging: integrating odometry leads to drift, loop closures are unavailable in previously unseen areas and embedded platforms provide limited computational power. 
We present a fully onboard UAV system developed for the SPRIN-D Funke Fully Autonomous Flight Challenge, which required 9~km long-range waypoint navigation below 25~m AGL (Above Ground Level) without GNSS or prior dense mapping. 
The system integrates perception, mapping, planning, and control with a lightweight drift-correction method that matches LiDAR-derived local heightmaps to a prior geodata heightmap via gradient-template matching and fuses the evidence with odometry in a clustered particle filter.
Deployed during the competition, the system executed kilometer-scale flights across urban, forest, and open-field terrain and reduced drift substantially relative to raw odometry, while running in real time on CPU-only hardware. 
We describe the system architecture, the localization pipeline, and the competition evaluation, and we report practical insights from field deployment that inform the design of GNSS-denied UAV autonomy.
\text{SUPLEMENTARY MATERIALS:}
\href{https://gnssdenied.github.io/}{https://gnssdenied.github.io/}
\end{abstract}
\vspace{-0.5em}


\section{INTRODUCTION}

\acp{UAV} are increasingly deployed for infrastructure inspection, logistics, and search-and-rescue. Many of these missions require long-range, low-altitude flight in areas where \acp{GNSS} are unreliable, unavailable, or denied. 
Without \ac{GNSS}, \acp{UAV} typically rely on visual–inertial or LiDAR odometry to navigate a previously unseen environment. 
These methods are consistent locally, but accumulate unbounded drift in previously unvisited areas, where loop closures cannot be used. 
When flying over kilometer-scale trajectories, this leads to position errors too large for waypoint-based navigation. 
The core problem addressed in our paper is therefore how to manage the drift using available geodata on a low-altitude resource-constrained UAV.

Existing methods address parts of the problem -- deep-learning geolocalization achieves high accuracy but is too computationally heavy to run in real-time on embedded platforms, while odometry-only systems based on \ac{VIO} or \ac{LIO} are efficient but accumulate excessive drift. 
At high altitudes, UAVs can use satellite data to correct their positions, but this becomes unfeasible close to the ground or under trees.
Fully onboard systems that scale to kilometer ranges in diverse unseen environments remain rare, since outdoor geodata-based drift correction that is both lightweight and robust has not been solved.
Addressing these challenges requires not only an individual algorithm but the integration of perception, localization, planning, and control into a reliable onboard system.  

We study this problem in the context of the \emph{SPRIN-D Funke Fully Autonomous Flight Challenge}, which provided a stringent test of GNSS-denied autonomy. 
In response, we developed a UAV system that integrates onboard mapping, planning, and mission execution with a lightweight geodata-based localization method for drift correction. 
Our method's key idea is to exploit heightmap gradients as a compact and robust structural signature---simple enough for real-time onboard use, yet distinctive enough to correct drift over kilometer-scale trajectories.
The system was deployed during the competition, where it successfully executed kilometer-scale flights in mixed environments and achieved substantial drift reduction compared to raw odometry.

\begin{figure}
  \centering
  \includegraphics[width=0.49\textwidth]{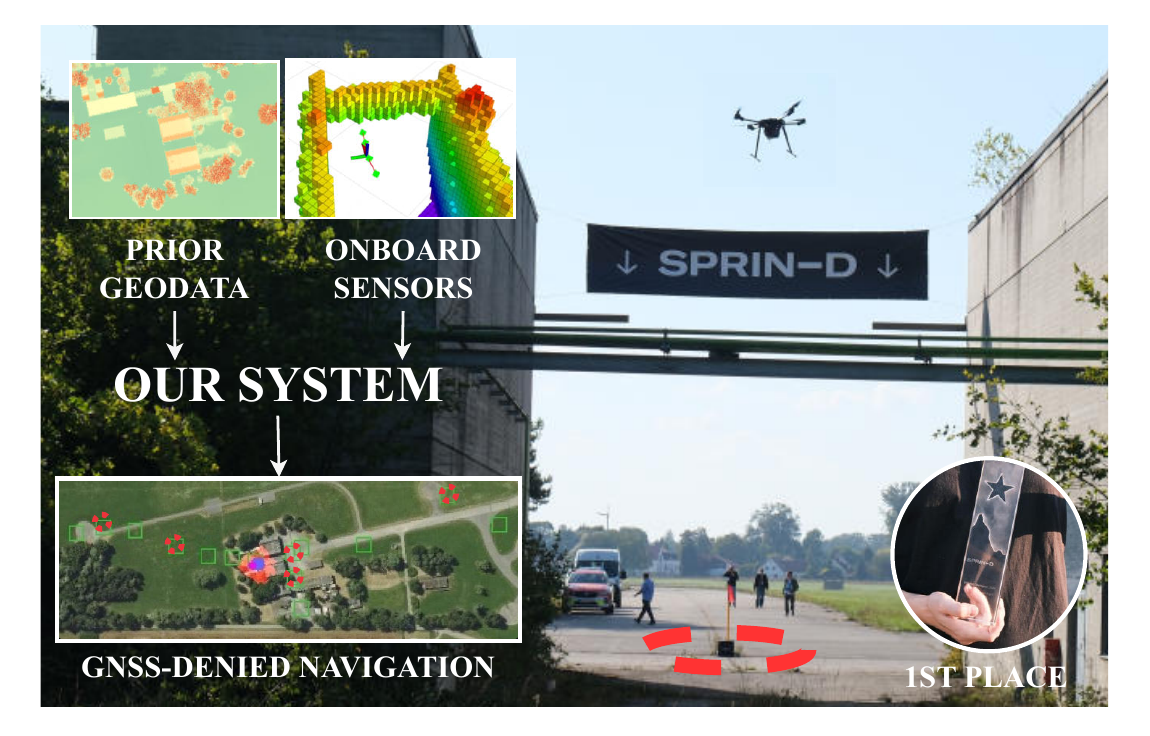} 
  \caption{The autonomous system for long-range GNSS-denied flight presented in this paper achieved 1st place in the international Fully Autonomous Flight Challenge.} 
  \label{fig:system_scheme}
  \vspace{-1em}
\end{figure}

\subsection{Problem definition}
The SPRIN-D Challenge required an autonomous \ac{UAV} with total mass below \SI{25}{\kilo\gram} to fly a prescribed \SI{9}{\kilo \meter} long sequence of waypoints, each waypoint marked by a red flag on a \SI{1}{\meter} pole, without relying on any GNSS-based localization. The waypoint positions were provided on a printed map with an uncertainty of approximately \SI{20}{\meter}. Throughout the mission, the UAV had to remain below \SI{25}{\meter} \ac{AGL} and autonomously avoid obstacles such as trees, buildings, or water curtains sprayed from a fire truck.

The competition environment combined urban areas, forests, and open fields. Although the area was announced beforehand, reconnaissance flights or custom dense mapping were explicitly prohibited, precluding the use of pre-recorded maps or ground-truth data. Together, these requirements defined not only a localization problem but a full autonomy challenge, demanding a UAV system capable of reliable long-range navigation and mission execution using prior data of the deployment area, and onboard sensing with constrained computation.
\subsection{Related work}
Localization without \ac{GNSS} is a widely studied problem in robotics, but remains challenging for long-range \ac{UAV} missions. 
Odometry methods such as \ac{VIO} and \ac{LIO} provide local motion estimates, yet accumulate drift that is usually corrected through loop closures. 
In real-world missions, e.g. search and rescue, \acp{UAV} operate in previously unseen environments where revisits are rare and loop closures cannot be relied upon. 
We tested state-of-the-art visual \ac{SLAM} systems such as RTAB-Map \cite{rtab_map} and ORB-SLAM3 \cite{orbslam3_tro} in a high-fidelity simulator \cite{capek_flightforge_2025}, and found that both performance degraded significantly in a long-range scenario (beyond \SI{1}{\kilo \meter}), as their memory and compute demands grow with the size of the environment. 
Moreover, RTAB-Map was unable to maintain quality odometry in faster flight speeds (beyond \SI{2}{\meter\per\second}), while ORB-SLAM3 suffered from tracking loss in textureless areas.
This confirmed that conventional \ac{SLAM} cannot serve as the basis for our system and motivates approaches that combine odometry with external geodata to maintain bounded error.

Several large-scale autonomy efforts, most notably the DARPA Subterranean Challenge \cite{darpa_overall}, demonstrated advanced GNSS-denied navigation capabilities. 
SubT is the closest system-level reference, but its setting was fundamentally different: multi-robot teams operating primarily in underground environments, often supported by larger platforms and heterogeneous sensing. 
In contrast, the SPRIN-D Challenge focused on \acp{UAV} to operate autonomously in large outdoor environments at altitudes constrained below \SI{25}{\meter}, without any prior mapping and without the support of other robots.

The absence of loop closures during the mission naturally leads to the problem of geolocalization within prior geodata.
Existing geolocalization methods can be broadly divided into camera-based, LiDAR-based, and semantic approaches. 
Camera-based methods often align aerial or UAV imagery with satellite maps. 
High-altitude matching \cite{osm_camera_global_aerial_localisation, goforth_aerial_image_matching} presents a viable solution in higher altitudes, but at low altitudes (below \SI{25}{\meter}) the viewpoint differs drastically, making roofs, facades, and vegetation inconsistent with satellite imagery. 
A range of works \cite{fervers_2022_continuous_selflocalization, wag, rewag, netvlad_particlefilter_image_geoloc,deallert_particle_filter, visionbased_geoloc_ground_cmu_2014} consider low-altitude or ground-based localization.
\cite{fervers_2022_continuous_selflocalization} 
combine a 3D lidar with camera data and train an end-to-end matching model for localizing a grounded agent with a forward-facing camera.
\cite{wag, rewag} train Siamese networks to match ground and satellite image embeddings.
The authors of \cite{netvlad_particlefilter_image_geoloc} train a cross-view-matching network for satellite-ground localization and combine the matches with odometry using a particle filter \cite{deallert_particle_filter}. 
The method in \cite{visionbased_geoloc_ground_cmu_2014} uses panoramic ground imagery warped to bird’s-eye view. 
While effective in their respective domains and on certain datasets, these methods assume ground agents or structured viewpoints and are not directly applicable to small UAVs without onboard GPU support deployed in cluttered outdoor mixed environments.

LiDAR-based methods use structural cues to improve robustness. Examples include learned place recognition fused with odometry \cite{lidar_localizing_faster_2020}, tree segmentation for forested areas matched with prior aerial scan of the area \cite{canopy_2023_ral}, or heightmap matching \cite{kaslin_heightmaps_2016}. However, these are typically restricted to small scales or specific environments, and are not easily transferable to urban–forest missions. Semantic approaches aim to mitigate appearance variability by extracting higher-level categories. Roads \cite{cmu_loc_vision_geoloc}, sematic maps \cite{christie_semantics_2016}, or canopy structures \cite{canopy_2023_ral} have been exploited as robust cues, often fused with odometry inside a particle filter. These methods offer seasonal robustness, but are limited in scope and domain.

Our work follows the general paradigm of combining a similarity estimation front-end with odometry in a particle filter \cite{deallert_particle_filter, thrun_particle_filter}, but adapts it to the unique constraints of the task we address: scalability for multi-kilometer flights, GNSS-denied operation, mixed urban–forest domains, and the requirement for reliable out-of-the-box functionality on test day. Unlike previous methods, we designed a new similarity estimation approach based on tall-object detection, which bridges LiDAR-based and semantic cues. 
Tall objects provide a stable and distinctive cue visible at low altitudes in both urban and forest environments, making them particularly suitable for the challenge setting.
This choice proved to be computationally efficient, robust across diverse large environments, and suitable for real-world deployment without extensive parameter tuning.

\subsection{Contributions}
We present a complete UAV system capable of kilometer-scale GNSS-denied flight below \SI{25}{\meter} AGL, operating fully onboard without a dedicated GPU acceleration.
Moreover, we introduce a novel drift-correction method based on template matching of LiDAR-derived heightmap gradients, fused with odometry in a clustered particle filter. 
Finally, we validated the system in urban, forest, and open-field environments during the SPRIN-D Challenge, where the organizers fixed the conditions of the challenge which led to an objective evaluation of all teams' solutions. 
As the only team, we achieved kilometer-scale flights in adverse conditions with substantial drift reduction relative to raw odometry, and we report practical lessons learned to guide future UAV design in GNSS-denied settings.

\section{SYSTEM FOR GNSS-DENIED AUTONOMOUS FLIGHT}
\begin{figure}[!h]
  \centering
  \includegraphics[width=0.5\textwidth,trim={0.6cm 0.2cm 0.5cm 0.2cm},clip]{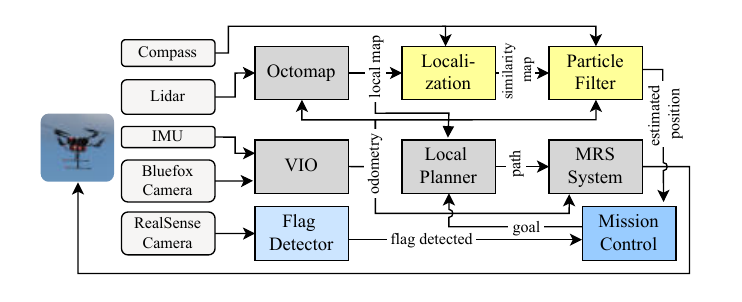}  \caption{System overview. Sensor inputs from LiDAR, cameras, IMU, and compass feed modules for visual–inertial odometry, obstacle mapping, and flag detection. A similarity-based localization module fuses odometry with prior geodata in a particle filter. Mission control coordinates planning and control for waypoint-based flight. }
  \label{fig:whole_system_scheme}
  \vspace{-1.5em}
\end{figure}

The task of GNSS-denied long-range navigation requires integration of perception, localization, planning, and control
into a reliable onboard system. The structure of the proposed solution is shown in \reffig{fig:whole_system_scheme}. 

\subsection{Hardware}
\begin{figure}[!h]
  \centering
  \includegraphics[width=0.5\textwidth,trim={1cm 0cm 0.5cm 0.2cm},clip]{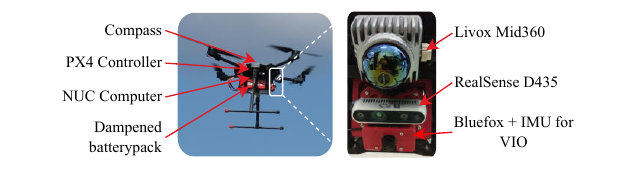} 
  \caption{Overview of the UAV platform and integrated sensor suite. The multirotor airframe carries a Livox Mid-360 LiDAR for mapping and localization, an Intel RealSense D435 depth camera for short-range obstacle perception, a Bluefox RGB camera and \ac{IMU} for \ac{VIO}, and an onboard Intel NUC computer for real-time processing.}
  \label{fig:hardware}
  \vspace{-1.5em}

\end{figure}

The UAV platform was based on \cite{hert_2023_jint}, and equipped with a heterogeneous sensor suite (\reffig{fig:hardware}) designed to enable robust state estimation, environment perception, and mission execution in GNSS-denied conditions. 
A Livox Mid-360 LiDAR sensor provided dense point clouds that served as the primary source for obstacle detection, local mapping, and heightmap generation, thereby forming the backbone of both planning and localization. 
Complementary visual sensing was provided by an Intel RealSense D435 depth camera for short-range obstacle perception, and a global-shutter Bluefox RGB camera paired with an inertial measurement unit (IMU) for visual–inertial odometry (VIO) using the OpenVINS framework \cite{openvins_2020}. 
To ensure the integrity of inertial measurements, the \ac{IMU} ICM42688 and \ac{VIO} camera Bluefox2 are mounted on a custom battery case that was mechanically decoupled from the airframe by 3D-printed silent blocks, effectively attenuating high-frequency vibrations induced by the propulsion system. 
An onboard magnetometer was employed to provide absolute heading measurements for global frame alignment, although its reliability was observed to degrade in magnetically disturbed environments, such as steel-reinforced concrete runways.
All computation was performed onboard by an Intel NUC i7 16~GB RAM computer, chosen for its balance of performance and weight, and critically without reliance on dedicated GPU acceleration. 
This hardware configuration enabled real-time execution of mapping, planning, localization, and mission control during fully autonomous flights.

\subsection{Visual Inertial Odometry}
\begin{figure}[!h]
  \centering
  \includegraphics[width=0.4\textwidth,trim={3cm 3cm 3cm 3cm},clip]{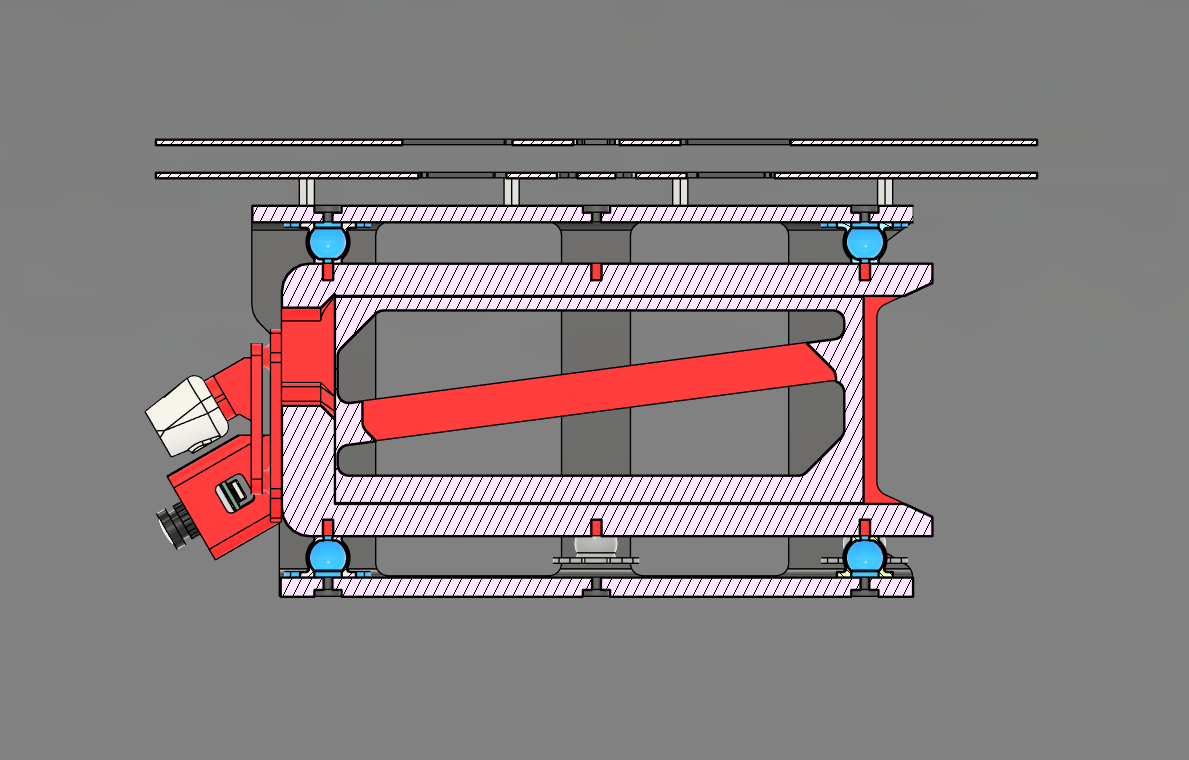} 
  \caption{Custom 3D-printed mount with integrated sensors, mechanically decoupled by silent blocks (blue) to dampen vibrations and improve \ac{VIO} robustness.}
  \label{fig:tlumeni}
  \vspace{-1.5em}
\end{figure}

Accurate state estimation is essential for feedback control and for providing a prior to the localization module. 
For this purpose, we employ \ac{VIO}, which offers a good balance between computational efficiency and robustness, making it well suited for real-time UAV operation in diverse outdoor environments.
We have opted to utilize monocular \ac{VIO} provided by OpenVINS \cite{openvins_2020}.  
We found that the reliability and practical usability of visual–inertial odometry critically depend on isolating the \ac{IMU} from high-frequency vibrations generated by motors and propellers, which would otherwise corrupt the inertial measurements and make the odometry unusable. 
Moreover, it is practical to mount the camera and \ac{IMU} as close to each other as possible.
To address this, both the \ac{VIO} camera and \ac{IMU} are mounted on the battery case, mechanically decoupled from the drone body using additively manufactured silent blocks, as shown in \reffig{fig:tlumeni}.  
The inherent mass of the battery further contributes to the damping effect, thereby reducing the transmission of vibrations. 

\subsection{Mapping, planning and feedback control}
Safe low-altitude flight in complex environments requires the ability to continuously perceive surrounding obstacles and to compute collision-free trajectories in real time. 
To this end, point clouds from the Livox Mid-360 LiDAR are incrementally integrated into a local occupancy map using the OctoMap framework \cite{mrs_darpa, octomap}.
The map is centered on the UAV body frame~$\mathcal{B}$, spans an area of $60 \times \SI{60}{\meter}$, and is updated at \SI{10}{\hertz}, thereby providing a compact yet sufficiently detailed representation of the immediate surroundings. 
Collision-free paths are found using the A* search algorithm on the euclidean signed distance field constructed from the local occupancy map. 
The selected path is subsequently transformed into a dynamically feasible trajectory using a polynomial trajectory generation module~\cite{richter2016polynomial}.
The trajectory is then tracked by the underlying model predictive reference tracking and control pipeline \cite{mrs_system}. 
In this way, the mapping and planning subsystem forms the essential link between perception and control, enabling safe navigation in cluttered GNSS-denied environments.
Further details of our mapping, planning and feedback control pipeline can be found in \cite{mrs_darpa}.

\begin{figure*}[htb]
  \centering
    \resizebox{0.95\textwidth}{!}{%
  \begin{tikzpicture}[
    auto,
    node distance=1.0cm and 2.5cm,
    every state/.style={
        rectangle,
        draw=black,
        text centered,
        align=center,
        line width=0.8pt,
         inner sep=10pt
    },
    thick_border/.style={
        line width=1.2pt
    },
    double_border/.style={
        double,
        double distance=1.5pt
    }
]

\node[state, thick_border, double_border] (S1) at (0, 0) {Prepare UAV, Take off};
\node[state, thick_border, below=of S1] (S3) {Wait for Mission Start};
\node[state, thick_border, double_border, right=of S3] (S4) {Waypoints Navigation};
\node[state, thick_border, above=of S4, yshift=1cm] (S_Detect) {Waypoint Detection};
\node[state, thick_border, right=of S_Detect] (S_ZigZag) {Search Pattern};
\node[state, thick_border, right=of S4] (S_Home) {Return Home};
\node[state, thick_border, double_border, right=of S_Home] (S_Land) {Land};

\draw[->] (S1) -- node[right, pos=0.5] {Success} (S3);
\draw[->] (S3) -- node[above] {Start mission} (S4);

\draw[->] (S4) -- node[above, sloped, pos=0.5] {Waypoint} node[below, sloped, pos=0.5] {reached}(S_Detect);

\draw[->] (S_Detect.west) to[out=180, in=180, looseness=0.7] node[left, pos=0.5] {Detection} (S4.west);

\draw[->] (S_Detect) -- node[above] {detection too far} node[below] {or no detection} (S_ZigZag);

\draw[->] (S_ZigZag.south west) -- node[above, sloped, pos=0.5] {search complete} node[below, sloped, pos=0.5] {or timeout}(S4.north east);

\draw[->] (S4) -- node[above] {all waypoints} node[below] {cleared}(S_Home);
\draw[->] (S_Home) -- node[above] {Home reached} (S_Land);

\draw[->, dashed] (S4) edge[bend right] node[below, sloped, pos=0.5] {returnHomeSrv} (S_Home);
\draw[->, dashed] (S_ZigZag) -- node[right, pos=0.5] {returnHomeSrv} (S_Home);

\end{tikzpicture}
  }
  \caption{The mission control state machine}
    \label{fig:state_machine}
  \vspace{-1.5em}
\end{figure*}
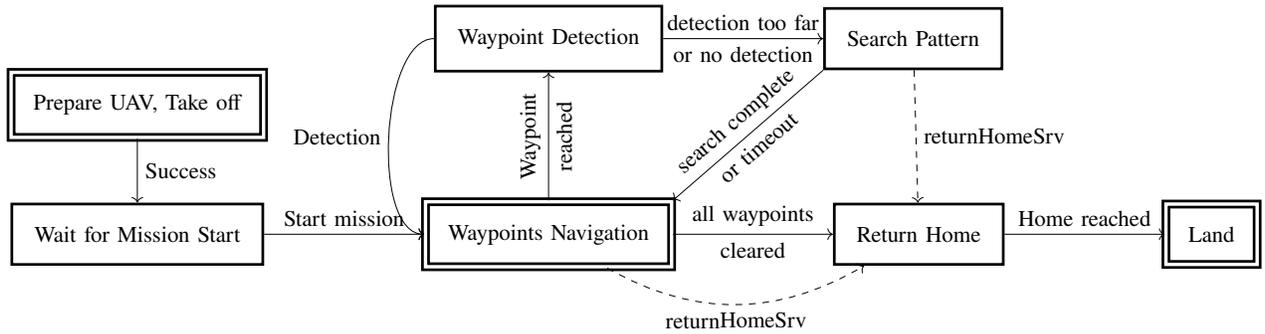

\subsection{Mission control}
The execution of the mission is governed by a finite-state machine shown in \reffig{fig:state_machine}.
The state machine starts by the preparation of the \ac{UAV} and the take-off procedure. 
Once the mission starts the system enters the navigation state and the \ac{UAV} starts to fly to the expected position of the first waypoint.
Once the \ac{UAV} enters a radius of \SI{15}{\meter} around the expected waypoint position, it switches to the waypoint detection state.
In the detection state the waypoint detection module is activated and images are being processed to detect the flags.
If the flag is detected, the system switches to the overflight state, where the \ac{UAV} flies over the waypoint to record the observation.
If the flag is not detected at the estimated location, the UAV executes a square search pattern centered on the waypoint. 
Detection at any time interrupts the search and triggers the overflight state; Failure to detect after the search concludes results in a transition to the next waypoint.
After reaching the last waypoint, the \ac{UAV} returns to the take-off position and lands.

\subsection{Digital twin driven development}
\begin{figure}[htb]

    \centering
    \begin{tikzpicture}[scale=0.7, transform shape]

        \node[inner sep=0] (image) at (0,0) {
          \includegraphics[width=0.7\textwidth]{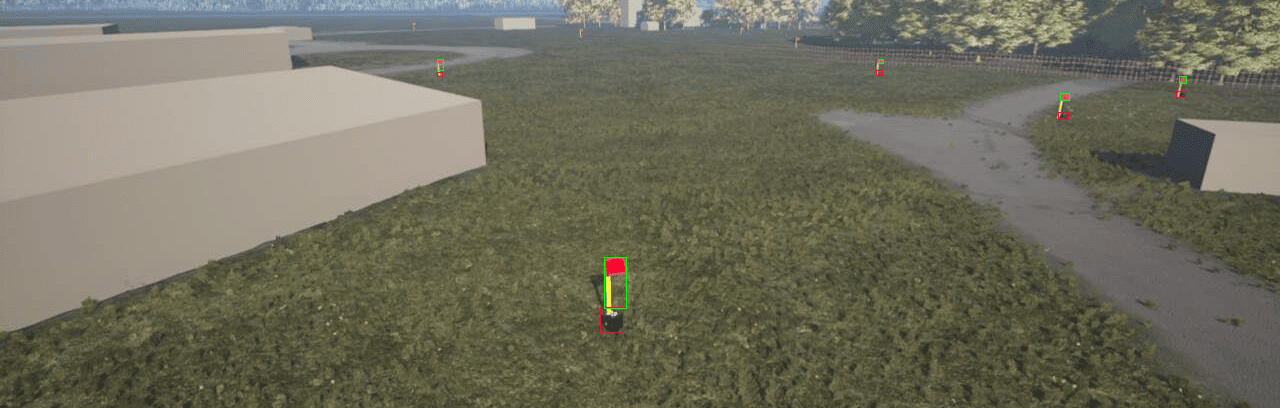}
        };

        \begin{scope}
            \clip (image.south west) -- (image.north east) -- (image.north west) -- cycle;

            \node[inner sep=0] at (0,0) {
                \includegraphics[width=0.7\textwidth]{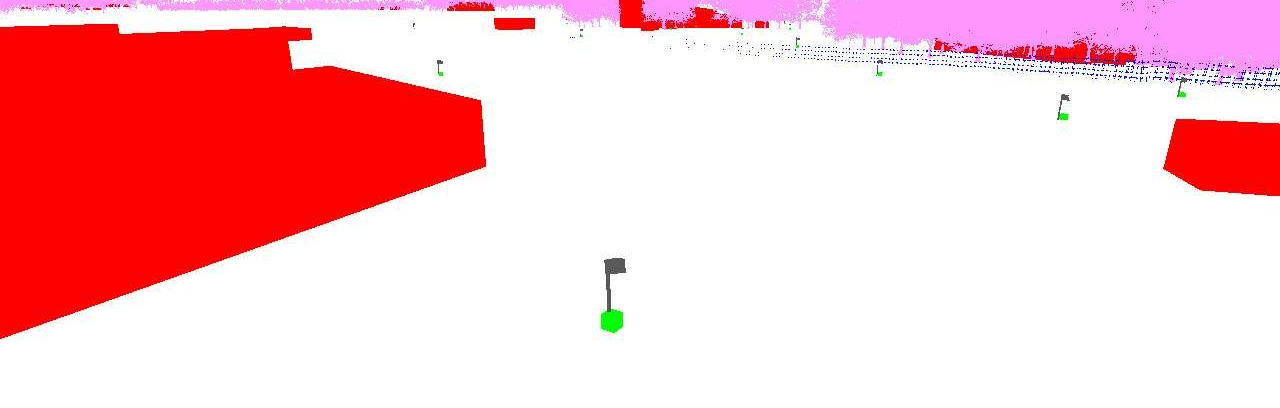}
            };
        \end{scope}

        \node[image_label, anchor=north west] at (image.north west) {(a)};

        \node[image_label, anchor=south east] at (image.south east) {(b)};

    \end{tikzpicture}
    \caption{Images from the digital twin environment: (a) segmentation mask, (b) RGB camera image with identified waypoints.}
    \label{fig:sim}
    \vspace{-1.5em}
\end{figure}
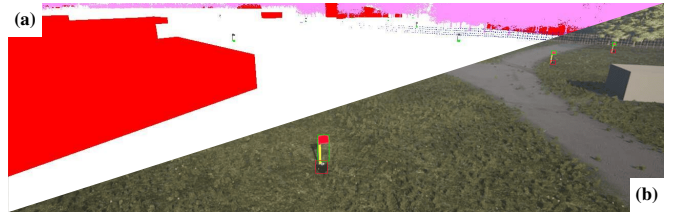

The robotic development process requires extensive testing and simulation of the individual components, as well as the whole system. 
For that purpose, we have created a digital twin of the environment in the FlightForge simulator \cite{capek_flightforge_2025} as similar as possible to the deployment area. 
The environment was created on the basis of multiple publicly available geodata \cite{geodaten_bayern_opendata}. The terrain was modeled based on \ac{DEM}, the vegetation (bushes, forest) was first identified in satellite images and then these 2D masks were used for the procedural generation of the corresponding vegetation. 
The same strategy was used for paved surfaces.
The buildings were modeled thanks to the building model (LoD2) available in the area from \cite{geodaten_bayern_opendata}.
The environment was extensively used for the development of the whole system, mainly the localization module as it is most strongly influenced by the environment.
We have gone through multiple iterations of the localization module, testing different approaches and parameters in the simulator before deploying it in the real world.
In addition to the localization, the simulated environment (\reffig{fig:sim}) was used to generate training data for the waypoint detector, as described in the next paragraph. 



\subsection{Waypoint detector}
The waypoint detection module was responsible for detecting the combination of a black box and a red flag on a yellow pole, as shown in \reffig{fig:sim}.
The module was based on the YOLOv8 architecture \cite{yolov8_ultralytics}.
First, a larger model YOLOv8m was trained on synthetic datasets generated in the FlightForge simulator \cite{capek_flightforge_2025}, where annotation are obtained automatically, as shown in \reffig{fig:sim}.
The synthetic dataset contained all the variations of the weather conditions, improving the generalization of the model.
After the initial training on synthetic data, the model was already able to detect the waypoints in real-world images on our own replica of the waypoint.
The pretrained model was then used to assist with annotation of real-world recordings containing the exact replicas of the waypoints, however the inference time was approximately \SI{300}{\milli\second}.
Finally, to reduce CPU load, a lightweight YOLOv8n model was trained on both synthetic and real data.
This smaller model was selected to meet onboard computational constraints while still providing reliable real-time detection with inference times of approximately \SI{100}{\milli\second} using CPU only.

\begin{figure}[!h]
  \centering
  \vspace{-0.5em}  \includegraphics[width=0.35\textwidth,trim={0.5cm 0.5cm 0.5cm 0.5cm},clip]{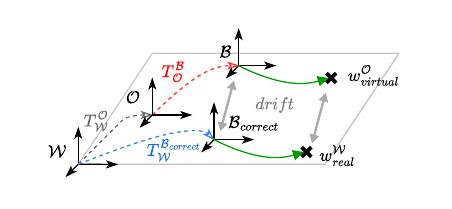}  \caption{Overview of the reference frames used in the system, where $\mathcal{W}$ is the georeferenced world frame, $\mathcal{O}$ is the odometry frame initialized on the known take-off position (i. e. known $T_{\mathcal{W}}^{\mathcal{O}}$), $\mathcal{B}$ is the body frame used by the feedback control and $\mathcal{B}_{correct}$ represents the corrected body frame (i.e. true position) in the world frame. The transform $T_{\mathcal{W}}^{\mathcal{B}_{correct}}$ is published by the localization system described in \refsec{sec:localization}. Given the position of the next waypoint $w_{real}^{\mathcal{W}}$, the virtual goal $w_{virtual}^{\mathcal{O}}$ is created to guide the \ac{UAV} to the correct position despite the odometry drift accumulated in $T_{\mathcal{O}}^{\mathcal{B}}$.}
  \label{fig:frames}
  \vspace{-1em}

\end{figure}

\section{ONBOARD LONG-RANGE GNSS-DENIED LOCALIZATION}
\label{sec:localization}

\begin{figure*}[!t]
  \centering
  \includegraphics[width=0.95\textwidth,trim={1cm 0cm 0.5cm 0.1cm},clip]{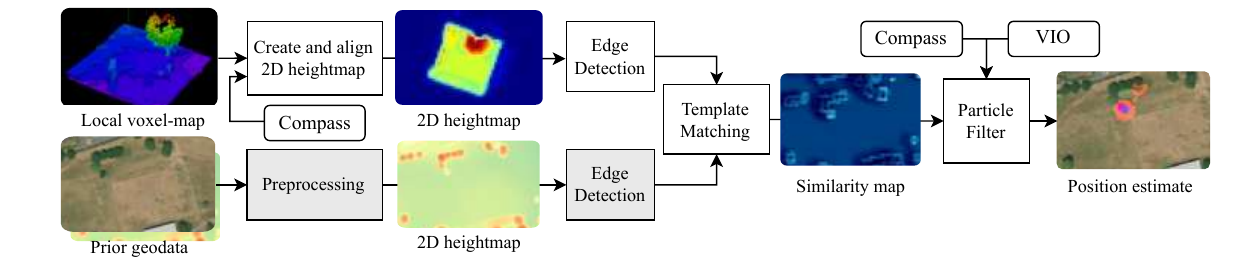}
  \caption{LiDAR-derived local heightmaps are registered to prior geodata heightmaps through gradient-based template matching. The resulting similarity map is subsequently integrated with odometry and compass measurements within a clustered particle filter to produce drift-corrected position estimates. Precomputed steps are denoted by grey boxes, whereas the remaining components are executed onboard.}
  \label{fig:localization_pipeline}
    \vspace{-1.4em}
\end{figure*}

This section presents the onboard localization pipeline for long-range GNSS-denied flight. 
A local heightmap, derived from the local occupancy map, is matched against preprocessed geodata, and the resulting similarity is fused with odometry in a clustered particle filter to provide robust position estimates. 
The pipeline described in this section is shown in \reffig{fig:localization_pipeline}.
\vspace{-0.7em}

\subsection{Heightmap Pre-Processing}

The prior data for the localization pipeline consist of a geo-referenced, north-aligned heightmap of the large-scale environment. 
For our experiments, we used a digital elevation model (\ac{DEM}) derived from open-source point clouds provided by~\cite{geodaten_bayern_opendata}. 
The point clouds were processed into a \ac{DEM} by estimating the ground surface and computing the relative height above terrain of all points using the LAStools software package~\cite{lastools}. 
Where high-accuracy \acp{DEM} are not available, obstacle heights above the terrain can alternatively be estimated from aerial RGB imagery using recent depth estimation models~\cite{depth_anything_v2, hrch} , as illustrated in \reffig{fig:localization_pipeline} and applied in our experiments (Flight ID 60 and 61), where accurate remote sensing geodata were not available. 
To ensure comparability, we apply the same preprocessing steps (discussed in the next section) to the prior DEM and to the local heightmaps generated onboard from LiDAR data. 
The heightmaps are constructed by calculating the max heights of the given pointcloud (from DEM or from the local map occupied cells) in \SI{1}{\meter} wide bins. This resolution allowed realtime operation at the multi-kilometer scale required by the competition while being precise enough for navigation.

The local heightmap itself is constructed from the online occupancy map. 
We assume that odometry drift remains bounded within the spatial extent of the local map, such that the resulting heightmap is not significantly distorted. 
In our experiments, the local map was configured as a square region with side lengths between \SI{30}{\meter} and \SI{60}{\meter}, depending on available computational resources. 
Finally, onboard compass measurements are used to align the heightmap with the north direction, ensuring consistency with the prior DEM, since the odometry frame $\mathcal{O}$ may be arbitrarily rotated relative to the Earth.

\subsection{Heightmap Gradient Matching}
Heightmaps can sometimes be matched using absolute elevation values, as demonstrated in prior work~\cite{kaslin_heightmaps_2016}. 
In our case, however, absolute height (e.g., \ac{AMSL}) is not reliably available on the \ac{UAV}, as barometric measurements are prone to high noise. 
Furthermore, the UAV might not see the ground at all times or the ground might be sloped, making it impossible to obtain absolute height using a ground plane model as in \cite{kaslin_heightmaps_2016}.
To address this, we perform matching based on \textit{gradients} of the heightmaps rather than absolute heights.
This eliminates the problem of vertical offsets between maps encountered in outdoor scenarios.

To further improve robustness, only gradients with an absolute value greater than \SI{5}{\meter} were considered during the SPRIN-D challenge runs. 
This filtering emphasizes tall, stable structures such as buildings and trees, while discarding small or transient objects (e.g., fences, vehicles, or bushes). 
The resulting binary edge maps mark strong gradients with ones, which are then compared to binary edge maps obtained from prior geodata using template matching.  
For the template matching function, we use the non-normalized (since the maps are normalized) correlation coefficient, defined as
\begin{equation}
 \resizebox{0.43\textwidth}{!}{
$R(x,y) = \sum_{x',y'} \left( T(x',y') - \bar{T} \right) \cdot \left( I(x+x',y+y') - \bar{I}_{x,y} \right)$},
\end{equation}
where $T$ denotes the local binarized heightmap, $I$ the prior map, $\bar{T}$ the mean of the local map, and $\bar{I}_{x,y}$ the mean of the matched patch at $(x,y)$. 
This metric was empirically found to outperform alternative functions (e.g., squared difference or normalized correlation), particularly when parts of the local heightmap were incomplete, as during straight-line flight or immediately after takeoff.  
Finally, because the heightmaps are computed at relatively coarse resolution, a Gaussian blur is applied to the resulting similarity map to reduce discretization artifacts.

\subsection{Particle filter}
Odometry and similarity maps are fused in a particle filter to provide a unified probability estimate of the UAV's position. 
While orientation is directly obtained from the compass, the particle filter maintains multiple hypotheses of the translational state. 
To enable real-time operation, resampling is triggered only after the UAV has traveled \SI{10}{\meter} according to odometry. 
Between resampling steps, particles are propagated by translating them according to the odometry estimate and aligning them with the compass heading.  

During the particle filter update, each particle is assigned a weight based on the normalized value of the similarity map at its projected location. 
These weights are then used as probabilities during resampling. 
When a particle is resampled, its new position is perturbed by Gaussian noise drawn from an empirically estimated odometry covariance, simulating the uncertainty propagation. 

Most prior works assume that particles form a single cluster and compute the position estimate as the mean of all particles. 
In practice, however, we frequently observed formation of multiple clusters caused by perceptual aliasing and odometry noise (see \reffig{fig:localization_pipeline}). 
To address this, we apply K-means clustering to the particle set and select the centroid of the largest cluster as the final position estimate.
The final global position estimate is used to calculate the direction towards the next waypoint.
To navigate towards the waypoint, we periodically set the goal position of the local planner ($w_{virtual}^{\mathcal{O}}$ in \reffig{fig:frames}) to be at a fixed distance from the UAV in the calculated direction to the next waypoint.
\vspace{-1.3em}

\section{EXPERIMENTS}
\vspace{-0.2em}
\begin{figure*}[htbp]
  \begin{subfigure}{0.32\textwidth}
    \centering
    \includegraphics[width=\linewidth,trim={0cm 1cm 0cm 5cm},clip]{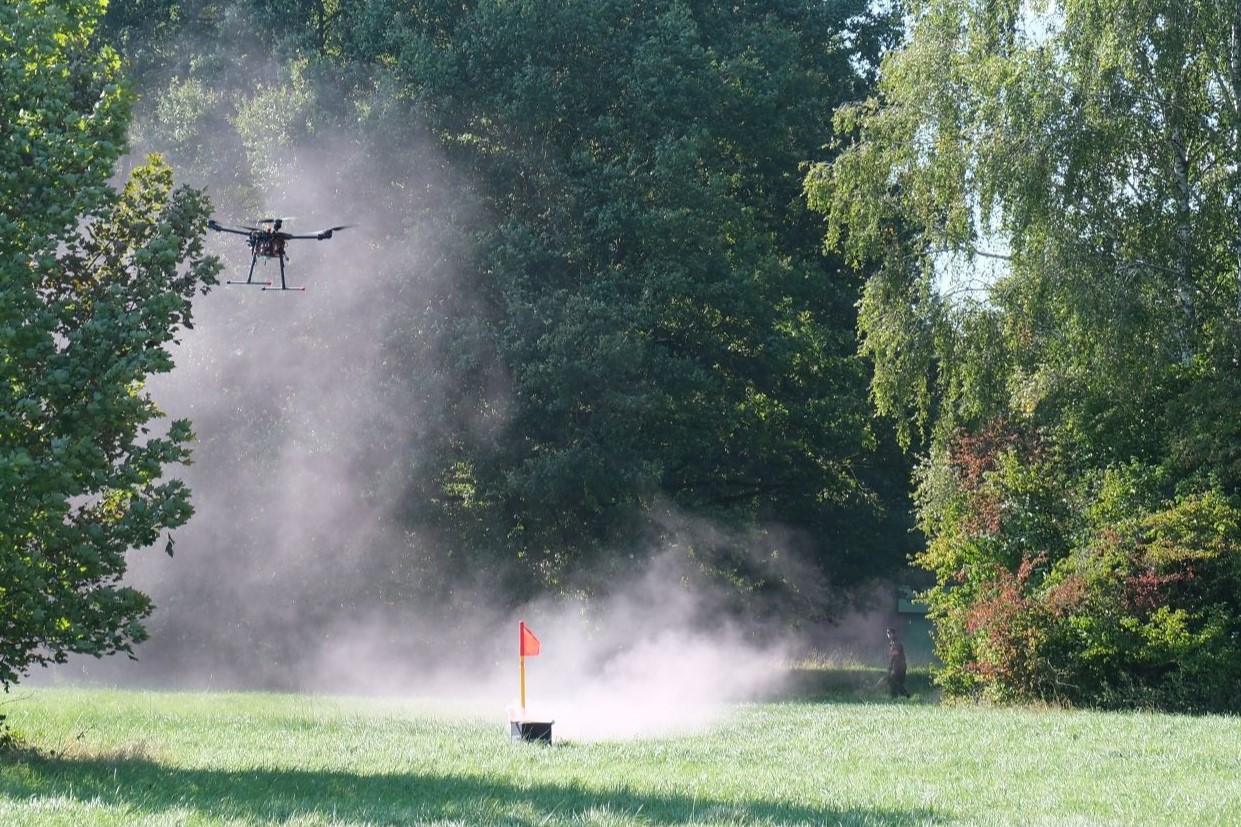}
    \caption{Forest environment (Flight ID 239)}
    \label{fig:forest}
  \end{subfigure}
    \hfill
      \begin{subfigure}{0.32\textwidth}
    \centering
    \includegraphics[width=\linewidth,trim={0cm 6cm 0cm 4cm},clip]{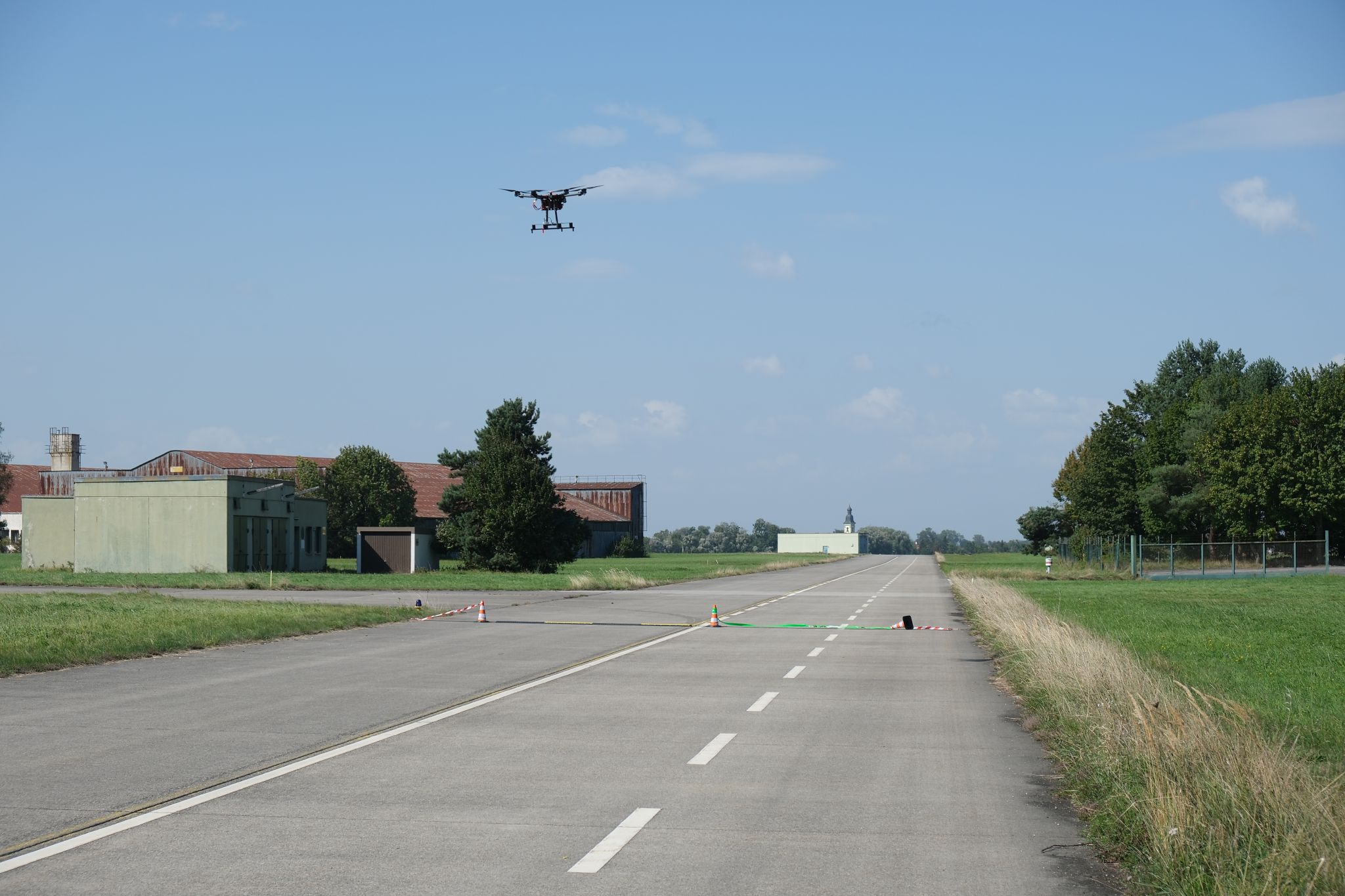}
    \caption{Runway/open field (Flight ID 208)}
    \label{fig:runway}
  \end{subfigure}
    \hfill
 \begin{subfigure}{0.32\textwidth}
    \centering
    \includegraphics[width=\linewidth,trim={0cm 4cm 0cm 6cm},clip]{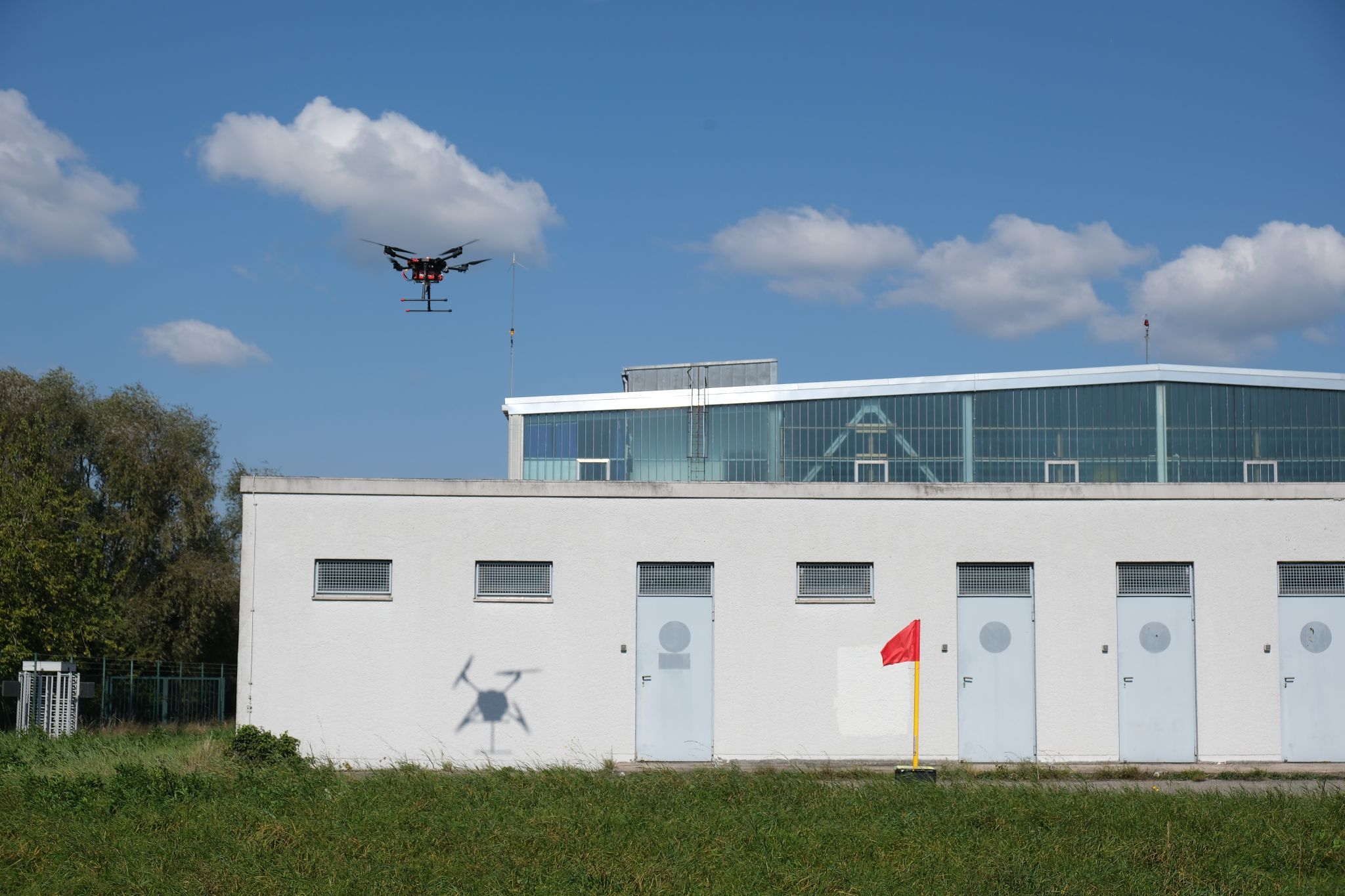}
    \caption{Urban environment (Flight ID 206)}
    \label{fig:urban}
  \end{subfigure}
  \begin{subfigure}{0.32\textwidth}
    \centering
    \includegraphics[width=\linewidth,trim={0cm 1.5cm 0cm 1.4cm},clip]{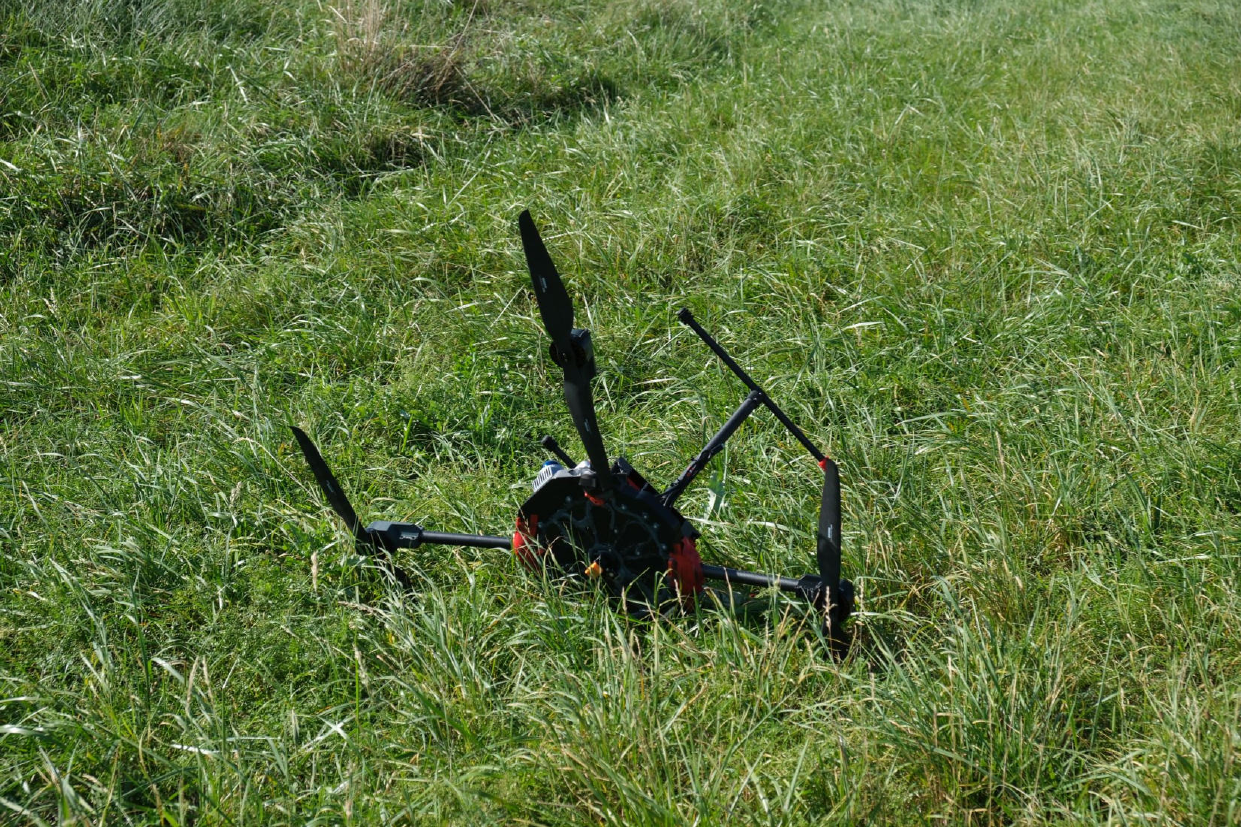}
    \caption{The result of a strong wind.}
    \label{fig:crash_no_people}
  \end{subfigure}
     \hfill
\begin{subfigure}{0.32\textwidth}
    \centering
    \includegraphics[width=\linewidth,trim={0cm 1.5cm 0cm 1.5cm},clip]{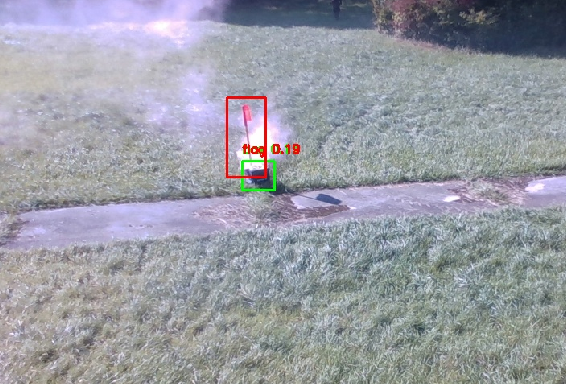}
    \caption{Successful onboard waypoint detection}
    \label{fig:flag_detection}
  \end{subfigure}
  \hfill
    \begin{subfigure}{0.32\textwidth}
    \centering
    \includegraphics[width=\linewidth,trim={0cm 0cm 0cm 0cm},clip]{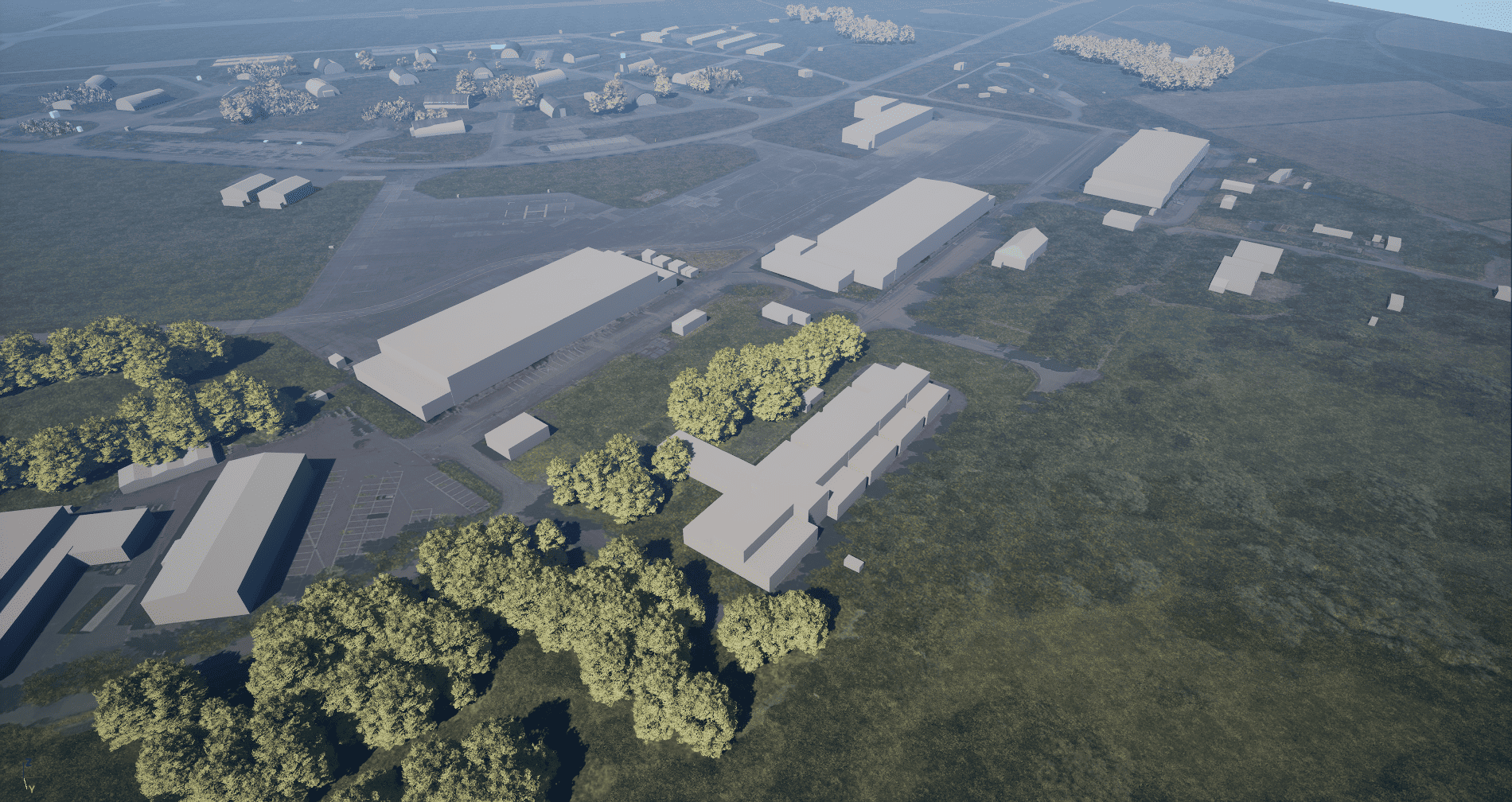}
    \caption{The competition area digital twin}
    \label{fig:erding_sim}
  \end{subfigure}
  \caption{Selected scenes from the system deployment, showing the three different competition environments, challenges we faced during the competition (strong wind, artificial smoke) and the simulated world used for the system validation and development.}
  \label{fig:scenes}
  \vspace{-1.4em}
\end{figure*}
The proposed robotic system was evaluated during the SPRIN-D Fully Autonomous Flight Challenge on the Erding airbase (Bavaria, Germany). 
The competition area covered a heterogeneous environment including (i) urban areas with buildings, fences, and individual trees, 
(ii) open fields, and 
(iii) semi-open forest sections. Each team was given two two-hour flight slots to validate their solutions under supervision of the jury.
On the final day of the competition, a \SI{9}{\km} course was defined by a sequence of waypoints distributed across all three environments (printed on a paper map distributed 30 minutes prior the mission). 
Each team had 2 hours to demonstrate the capability of their system and visit as many waypoints as possible.

\subsection{Evaluation Methodology}
\label{sec:gt}
Since no GNSS-based localization was permitted, accurate ground truth of the UAV position was not directly available in the challenge.
To estimate the real trajectory of the \ac{UAV}, we manually estimated the approximate ground truth trajectory of the \ac{UAV} from the \ac{VIO} data corrected by the camera/lidar footage from distinctive places and objects (where the real position of the \ac{UAV} is clearly identifiable with precision of $0$--$5$~m). 
We also present data from two test flights prior the competition, where \ac{GNSS} localization was available.
However, it must be noted that the system should not be evaluated only on the basis of the \ac{RMSE} that our localization achieved, but rather as a system that was able to perform multiple kilometer-scale flights in challenging conditions while being able to successfully find the target waypoints.

\subsection{Autonomous GNSS-denied flights}

\begin{figure}[!h]
  \centering
  \includegraphics[width=0.46\textwidth,trim={0cm 1cm 0cm 2.5cm},clip]{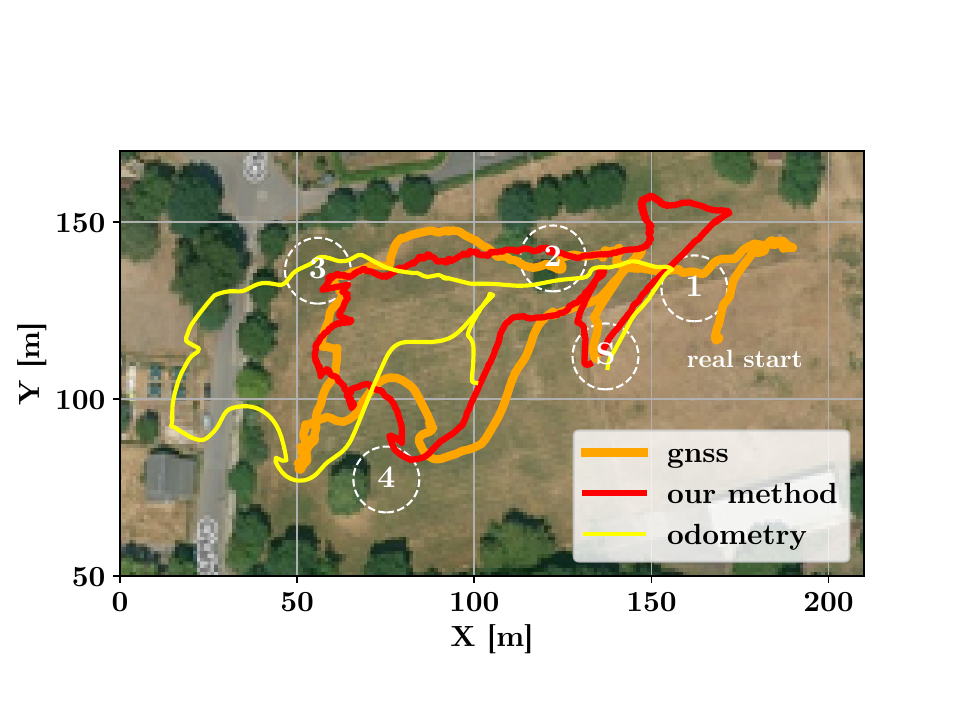}
  \centering
  \caption{Flight ID 61. The starting (and landing) position is marked with S and the waypoints are numbered in the respective order. During this experiment, the localization method was initialized (waypoint S) \SI{32}{\meter} off the starting position (see "real start" in the map). Despite the initial error, the localization method successfully relocalized by the end of the test flight.} 
  \label{fig:flight_61}
\end{figure}

\begin{figure}[!h]
  \centering
  \includegraphics[width=0.45\textwidth,trim={0.5cm 0.5cm -1cm 0.5cm},clip]{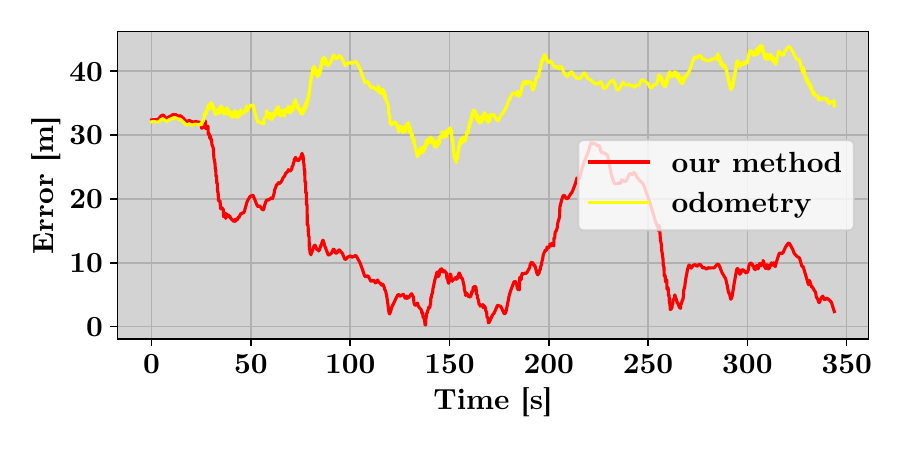}
  \caption{Flight ID 61. The localization error error of the proposed localization method and odometry compared to \ac{GNSS} ground truth.}
  \label{fig:flight_61_rmse}
  \vspace{-1em}
\end{figure}

The experiments performed before and during the competition are shown in \reftab{tab:results}.
Test flights 60 and 61 were conducted outside the competition area and thus have accurate GNSS ground truth. 
To test the recoverability and drift correction of our localization method, the localization module was intentionally initialized approx. \SI{32}{\meter} away from the true position in experiment number 61 shown in \reffig{fig:flight_61}.
In this experiment, our method managed to correct the position estimate when the UAV observed the trees and in the end reduced its error to approx. \SI{4}{\meter}, as shown in \reffig{fig:flight_61_rmse}.

\begin{figure*}[htbp]
  \begin{subfigure}{0.32\textwidth}
   \centering
  \includegraphics[width=\textwidth,trim={1cm 0cm 1cm 3cm},clip]{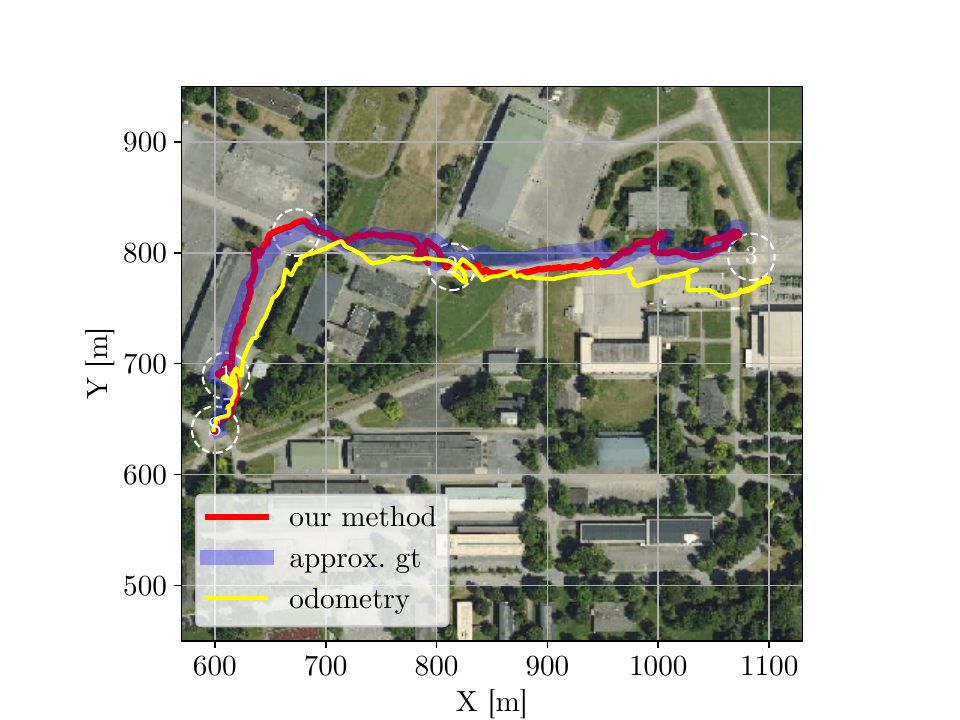}
  \caption{Flight ID 206}
  \label{fig:206}
  \end{subfigure}
     \hfill
\begin{subfigure}{0.28\textwidth}
     \centering
  \includegraphics[width=\textwidth,trim={1.6cm 0cm 2cm 1.2cm},clip]{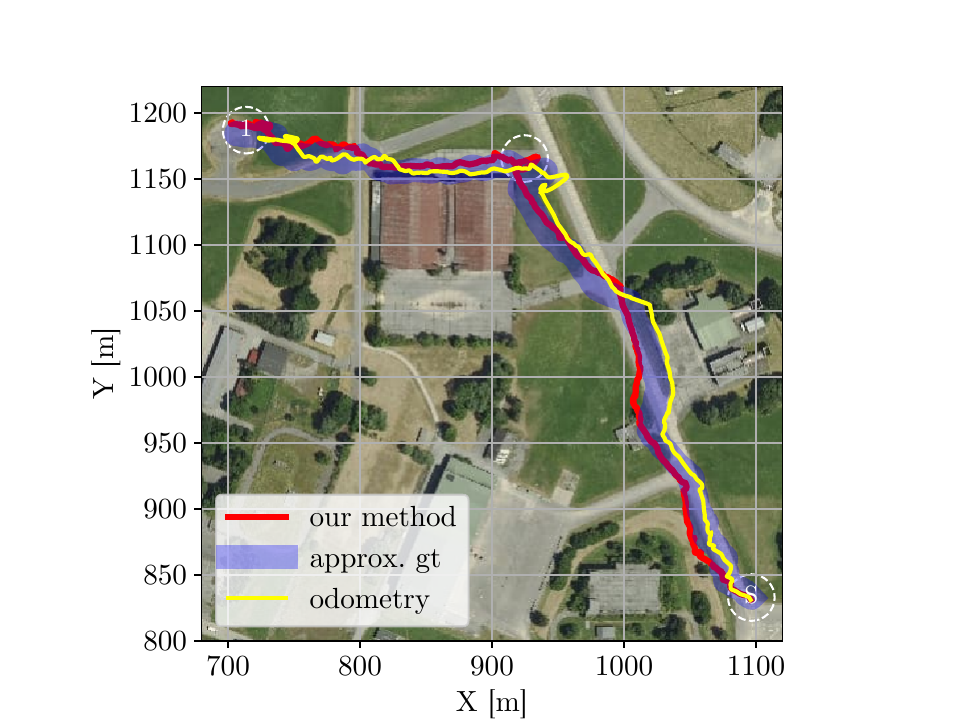}
  \caption{Flight ID 208}
  \label{fig:flight_208}
  \end{subfigure}
  \hfill
   \begin{subfigure}{0.31\textwidth}
   \centering
  \includegraphics[width=\textwidth,trim={0.5cm 0cm 1.5cm 1.3cm},clip]{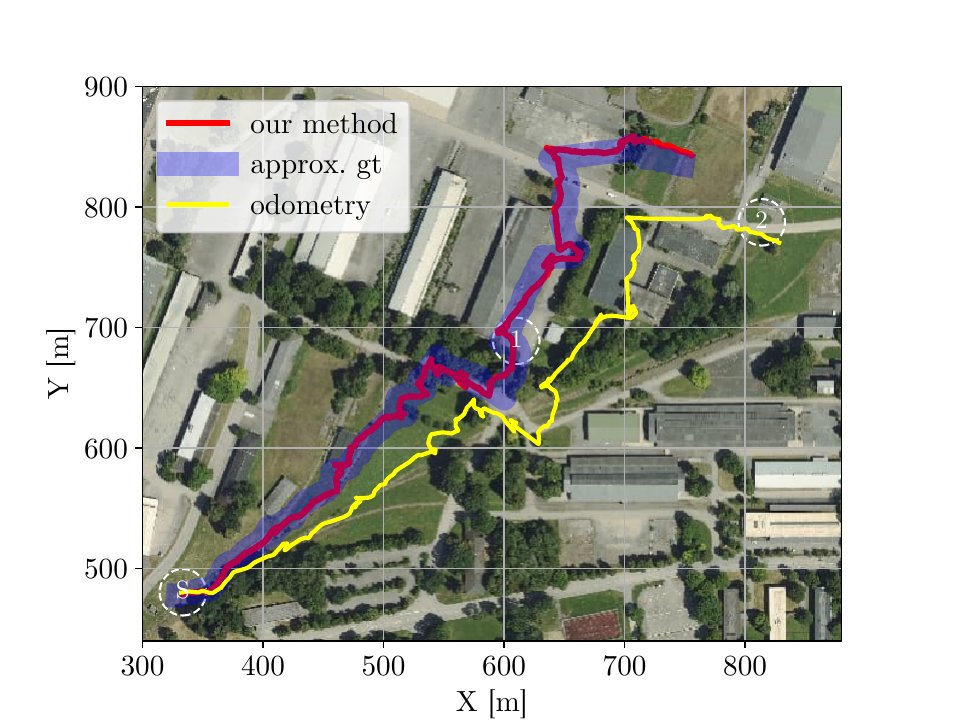}
  \caption{Flight ID 232}
  \label{fig:flight_232}
  \end{subfigure}
  \caption{Comparison of flight trajectories during the SPRIN-D competition. The results highlight the reduced drift of our method compared to raw odometry across different environments.}
  \label{fig:erding_plots}
  \vspace{-1.6em}
\end{figure*}

During the actual competition, our system was able to autonomously perform multiple kilometer-scale flights with overall localization \ac{RMSE} below \SI{11}{\meter} while the compass-aligned odometry had \ac{RMSE} of up to \SI{53}{\meter}.
Selected flights are shown in \ref{fig:erding_plots}.
Flights exceeding 1 km were consistently terminated due to battery constraints, rather than localization drift, indicating hardware limitations as the primary bottleneck.
The main reason was typically drained battery or a hardware issue, such as faulty WiFi card that restarted the onboard computer during the flight shown in \reffig{fig:flight_232}.
Despite the lack of accurate ground truth, it is possible to conclude that the proposed drift correction method significantly improved the localization accuracy over the kilometer-scale distances.
Our method showed better performance in urban environment, where there were more features and distinguishable objects that helped with the drift correction. 
In open field areas, the system mostly relied on the odometry.
The odometry from \ac{VIO} was fused with magnetometer measurements that helped to keep the odometry frame correctly oriented in the world coordinate frame.
In some areas of the competition, the compass provided incorrect data in the form of a slowly changing bias (as we can see right after the start in \reffig{fig:flight_232}) with an error of up to 30 deg. 
However, the proposed system was able to correct the magnetometer drift (when enough distinctive features were present).
\vspace{-1em}

\begin{table*}[htbp]
  \centering
  \begin{tabular}{
    @{}l
    S[table-format = 4.0]   
    S[table-format = 3]   
    S[table-format = 1]   
    S[table-format = 1]   
    c                       
    c
    l@{}                    
  }
    \toprule
    {Flight ID} &
    {Length [m]} &
    {Time [s]} &
    {RMSE\textsubscript{odom} [m]} &
    {RMSE\textsubscript{method} [m]} &
    {Waypoints detected} &
    {Area} &
    {Termination} \\
    \midrule
    60$^{*}$ & 830 & 606 & 8 & \bf{5} & -  & forest/open field & test finished \\ 
    61$^{*}$ &  495 & 343 & 36 & \bf{13} & - & forest/open field & test finished \\ 
    205$^{\dagger}$ &  939 & 622 & 31 & \bf{10} & 1/1  & urban & test finished \\ 
    206$^{\dagger}$ & 1023  & 621 & 27 & \bf{6} & 2/2 & urban & SW issue \\ 
    208$^{\dagger}$ & 872 & 500 & 14 & \bf{11} & 1/1  & urban/open field & low battery \\ 
    232$^{\dagger}$ & 1256 & 1021 & 53 & \bf{7} & 1/2  & forest & failsafe - HW issue \\ 
    239$^{\dagger}$ & 1371 & 977 & \bf{8} & 9 & 4/4  & forest & low battery \\ 
    242$^{\dagger}$ & 939 & 622  &  \bf{9} & 11 & 2/3  & urban/open field & low battery\\ 
    \bottomrule 
  \end{tabular}
  
    \caption{Summary of the real-world experiments. The a) ID of the flight, b) flight time, c) flight length, d) \ac{RMSE} of the odometry and our proposed method compared to GNSS ground truth($*$)/approximate ground truth described in \refsec{sec:gt} ($\dagger$) (both rounded to meters), e) number of waypoints with flag sucessfully detected, f) type of the environment and g) termination of the mission are reported.}
    \label{tab:results}
  \vspace{-1.7em}
\end{table*}


\subsection{Results of the competition}
A total of nine teams from European countries participated in the challenge.
The most significant difficulty proved to be establishing reliable odometry onboard the UAV, which is a prerequisite for stable autonomous flight. 
In addition, achieving accurate GNSS-denied geolocalization at low altitudes proved to be inherently difficult. Some teams used RGB satellite image-based matching, but this has proved  to be highly unreliable at such low altitudes.
As a result, only us and 2 other teams managed to realize autonomous flights under these demanding conditions for more than \SI{100}{\meter}.
The two other teams accumulated critical drift and were not able to reach more than 2 waypoints.
Our team succeeded in this task, ultimately achieving the best overall performance and being awarded first place in the competition.

\subsection{Lessons learned}
The challenge provided valuable insights into real-world GNSS-denied UAV deployment. 
Mechanical decoupling of the IMU and VIO camera from the drone frame proved essential: when the dampening mechanism failed, VIO drift increased sharply, showing the strong impact of parasitic vibrations. 
Although the compass appears a natural choice for global alignment, we confirmed that magnetometers are unreliable near buildings and reinforced concrete; however, our method was able to correct this drift when sufficient environmental features were available. More generally, this highlights the broader lesson in robotics that no single sensor can be fully relied upon, making sensor fusion essential for robust autonomy.
A general lesson is that overall system performance is constrained by its weakest component. 
In our case, the limited size of the local map restricted flight speed, which in turn capped the effective mission range despite accurate localization. 
Equally important in time-critical scenarios is rapid redeployment: the system must support fast, structured diagnostics of its components to localize the weak points of the system within minutes.

\section{CONCLUSIONS}

We have presented and successfully demonstrated a fully onboard system for reliable long-range UAV navigation in GNSS-denied environments. Our approach corrects odometry drift by aligning locally generated LiDAR heightmaps with prior geodata using a gradient-matching technique within a clustered particle filter, proven robust across varied terrains including urban, forest, and open fields. The system's performance in the SPRIN-D Challenge, where it achieved first place with RMSE below 11 m over kilometer-scale flights, validates its capability for real-world autonomous operation on CPU-only hardware. Crucially, our results demonstrate that for long-range missions, the ability to recover from periods of high uncertainty and re-localize is more critical than maintaining consistently low instantaneous RMSE. While challenges in endurance and operation in low-feature environments remain, this work provides a foundational blueprint for developing resilient GNSS-denied autonomy systems required for practical deployment.



\section*{ACKNOWLEDGMENT}
\small This work was funded by the German Federal Agency for Disruptive Innovation (SPRIN-D),
 by the Czech Science Foundation (GAČR) under research project no. 25-17779M,
by the European Union under the project Robotics and advanced industrial production (reg. no. CZ.02.01.01/00/22\_008/0004590) and by CTU grant no SGS23/177/OHK3/3T/13.


\bibliographystyle{IEEEtran}
\bibliography{root}

\end{document}